\documentclass[letterpaper, 10 pt, conference]{ieeeconf}  %

\IEEEoverridecommandlockouts                              %

\usepackage{graphics} %
\usepackage{epsfig} %
\usepackage{mathptmx} %
\usepackage{times} %
\usepackage{amsmath} %
\usepackage{amssymb}  %

\usepackage{booktabs}
\usepackage{xspace}

\usepackage{microtype}

\usepackage[pagebackref=true,breaklinks=true,colorlinks,bookmarks=false]{hyperref}

\newcommand*{\eg}{e.g.\@\xspace}
\newcommand*{\etal}{et\,al.\@\xspace}
\newcommand{\squeezeup}{\vspace{-4mm}}

\title{\LARGE \bf
LiDAR Meta Depth Completion
}

\author{Wolfgang Boettcher$^{1}$, Lukas Hoyer$^{1}$, Ozan Unal$^{1}$, Ke Li$^{1}$ and Dengxin Dai$^{2}$%
\thanks{$^{1}$ETH Zurich,  Switzerland
        {\tt\small wboettcher@student.ethz.ch}, 
        {\tt\small \{lhoyer, ozan.unal, ke.li\}@vision.ee.ethz.ch},}%
\thanks{$^{2}$ Huawei Technologies, Zurich Research Center
        {\tt\small dengxin.dai@huawei.com}}%
}

\begin{document}

\maketitle
\thispagestyle{empty}
\pagestyle{empty}

\begin{abstract}
    Depth estimation is one of the essential tasks to be addressed when creating mobile autonomous systems. While monocular depth estimation methods have improved in recent times, depth completion provides more accurate and reliable depth maps by additionally using sparse depth information from other sensors such as LiDAR. However, current methods are specifically trained for a single LiDAR sensor. As the scanning pattern differs between sensors, every new sensor would require re-training a specialized depth completion model, which is computationally inefficient and not flexible. 
    Therefore, we propose to dynamically adapt the depth completion model to the used sensor type enabling LiDAR adaptive depth completion. Specifically, we propose a meta depth completion network that uses data patterns derived from the data to learn a task network to alter weights of the main depth completion network to solve a given depth completion task effectively. 
    The method demonstrates a strong capability to work on multiple LiDAR scanning patterns and can also generalize to scanning patterns that are unseen during training. 
    While using a single model, our method yields significantly better results than a non-adaptive baseline trained on different LiDAR patterns. It outperforms LiDAR-specific expert models for very sparse cases.
    These advantages allow flexible deployment of a single depth completion model on different sensors, which could also prove valuable to process the input of nascent LiDAR technology with adaptive instead of fixed scanning patterns. The source code is available at \href{https://github.com/wbkit/ResLAN}{\UrlFont{github.com/wbkit/ResLAN}}
\end{abstract}

\section{Introduction}
    \begin{figure}
	    \centering
	    \includegraphics[width=0.9\columnwidth]{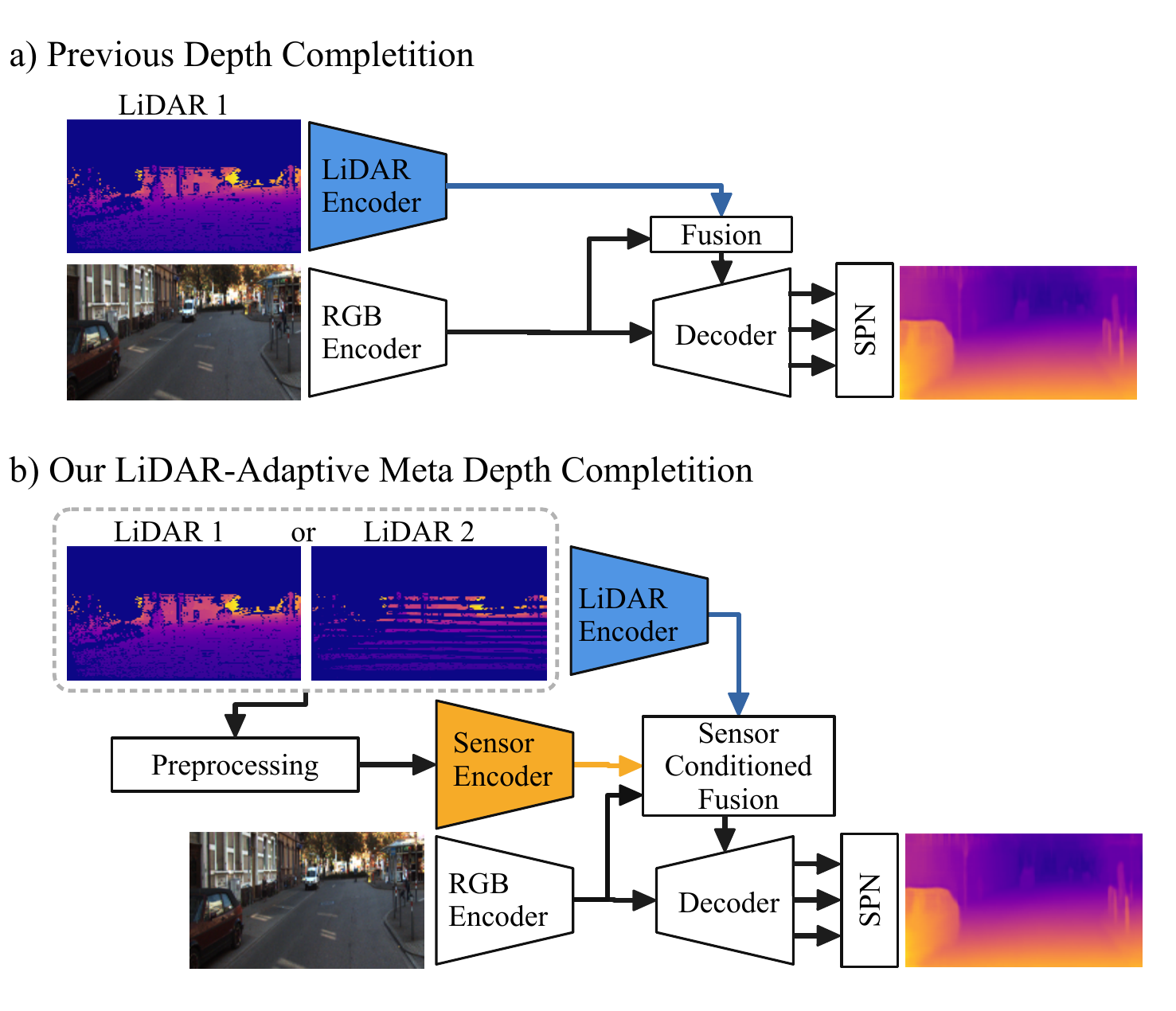}
	    \caption{\textbf{LiDAR adaptive meta depth completion}. (a) Previous depth completion methods such as SAN\,\cite{guizilini_sparse_nodate} are trained for a certain LiDAR sensor. If another sensor is used, they have to be re-trained. (b) In contrast, our method dynamically adapts the fusion to the used LiDAR input pattern. In that way, a single model can handle different LiDAR sensors.}
	    \label{fig:overview}
	\end{figure}
	Reliable and accurate depth prediction constitutes an essential computer vision capability needed to successfully execute a wide range of robotic tasks. Its use is especially prevalent for autonomously operating vehicles such as self-driving cars. Related tasks such as environment perception, navigation, and collision avoidance all require depth information. While RGB cameras are cheap, depth estimation based on single RGB images is very challenging \cite{wang_learning_2017, zhou_unsupervised_2017}. Therefore, there is strong research interest 
	 in how to supplement these methods with depth data from active sensors such as LiDARs\,\cite{uhrig_sparsity_2017, van_gansbeke_sparse_2019, ma_self-supervised_2019, guizilini_multi-frame_2022, liu_graphcspn_2022} or Radar\,\cite{long_radar-camera_2021, radar:depth:20,lo_depth_2021}. These active sensors generate direct but sparse depth measurements while cameras record dense RGB values. Hence, combining both can make the best of the two worlds. 
	 Depth completion models combine both sensor modalities and achieve remarkable performance for dense depth prediction. Using LiDARs, however, comes with a significant caveat that has not yet received due attention. Different LiDARs produce varying scanning patterns, which can cause significant performance degradation when a trained depth completion model is applied to another LiDAR. Thus, the LiDARs contribute notably to the domain gap between different datasets in a depth completion setting requiring a model to be retrained if changes to the sparse depth distribution are made. This issue poses a problem if, for instance, a depth completion model is created for a specific type of LiDAR, but multiple types of LiDARs may be used depending on the market cost or the user's applications during deployment. Additionally, recent LiDAR development has given rise to sensors with flexible scan patterns that can be adapted in real-time, like the Bajara Spectrum-Scan\,\cite{noauthor_bajara_nodate}, which also requires adaptive depth completion methods.
		
	To tackle this problem, we conceive a depth completion model that is structured to operate well on multiple depth input distributions without the need to retrain new models. To achieve that, we propose a meta depth completion network based on the Sparse Auxiliary Network Structure\,(SAN)\,\cite{guizilini_sparse_nodate}. We propose a novel extension to this network that enables LiDAR adaptive depth completion for the first time. The meta depth completion network contains a sensor encoder network, which learns to alter the weights of the main depth completion network to perform depth completion conditioned on the used sensor. It derives its input directly from the LiDAR sampling pattern, which enables it to generalize even to unseen data distributions e.g.\,LiDAR types.
	
	To train and evaluate our method, we simulate data of different types of LiDARs by filtering the data of a high-end LiDAR sensor such as a 64-beam LiDAR. This approach allows us to use the same dataset and avoid introducing any other types of domain gaps.
		
	Our main contributions summarize as follows:
	\begin{itemize}
	    \item To the best of our knowledge, we are the first to study LiDAR-adaptive meta depth completion. While previous methods require training a separate model for each used LiDAR, our novel approach dynamically adapts the model to the used scanning pattern of different LiDARs. This allows flexible deployment of a single depth completion model on different sensors.
        \item Specifically, we propose a novel LiDAR adaptive depth completion method that can perform depth prediction as well as completion for multiple types of LiDAR scanning patterns. This is achieved by introducing weight adaption networks that dynamically control the weighting of depth and image features when fusing both together.
        \item The proposed architecture significantly outperforms a non-adaptive baseline trained on different LiDAR patterns. Further, it even outperforms a specialized depth completion baseline, which is trained for a specific scanning pattern in the case of very sparse depth input. Our method is further capable of generalizing to previously unseen depth sample distributions.
    \end{itemize}

\section{Related Work} \label{sec:related}
    \paragraph{LiDAR Processing}
        Depth data obtained from LiDAR sensors distinguishes itself by its irregular sampling and sparsity by measurement principle. It generates a set of points in 3D space. In contrast, a camera records dense 2D data in a cartesian coordinate system but does not possess directly accessible depth information. For depth completion, the point cloud is usually projected onto a 2D plane allowing it to be processed akin to regular image data when using LiDAR data for creating depth maps\,\cite{chen_estimating_2018, guizilini_sparse_nodate, van_gansbeke_sparse_2019, lin_dynamic_2022}. Alternatively, the 3D nature of the data can be retained and specialised processing methods\,\cite{charles_pointnet_2017} be applied, which is more common for segmentation or object detection tasks\,\cite{shi_pointrcnn_2019}.

     \paragraph{LiDAR Depth Completion}
    	Depth completion, like depth prediction, aims to provide a model that generates a depth map from its inputs. In the latter case, only image data is available, making the setup simple and low-cost. Models learn to predict depth from an input image for every pixel inside it. This may involve fully supervised training \,\cite{laina_deeper_2016, eigen_depth_2014} or self-supervised methods \,\cite{ kok_review_2020, zhang_consistent_2021, patil_dont_2020}. In any case, the predictions of these models can only be correct up to a scale factor since depth prediction with one camera is an ill-posed problem. Thus, additional input data \eg LiDAR is required for accurate predictions leading to the adoption of depth completion networks. These models are usually designed as encoder-decoder networks where both inputs are firstly processed separately and then projected onto a common latent space from which they are processed in one stream. Ma and Karaman\,\cite{ma_sparse--dense_2018} note that, generally, the model performance for completion tasks is contingent on how well the models can treat cross-modality information as some of the early works \cite{kuga_multi-task_2017} did not manage to notably outperform depth prediction models when compared against each other. 
    	Besides the main combination of RGB images and LiDAR, some strategies further utilise semantic relations\,\cite{nazir_semattnet_2022}, or surface-normals\,\cite{xu_depth_2019}.

    \paragraph{LiDAR Adaption}
    Adapting depth completion networks to different LiDARs relates to the field of domain adaptation, with a change in LiDAR point sampling considered a domain change. A wide range of approaches for (unsupervised) domain adaptation has been developed during the past years\,\cite{yi_complete_2021, zhao_epointda_2021}. So far, concepts for depth estimation/completion have used style-transfer approaches on 2D data\,\cite{atapour-abarghouei_real-time_2018, zheng_t2net_2018}, and sparse-to-dense methods using synthetic data\,\cite{qiu_deeplidar_2019-1, atapour-abarghouei_complete_2019}. A different approach with a similar goal is pursued by Lopez-Rodriguez \etal\cite{lopez-rodriguez_project_2020}, who generate depth measurements from a scene simulation and use real LiDAR inputs from the target domain as a mask to sample depth measurements adapted to the target domain. RGB images are generated using a CycleGAN\,\cite{zhu_unpaired_2020}. Their findings give testament to the importance of the LiDAR sampling pattern when doing depth completion.
    While domain adaptation methods allow adjusting depth completion networks to different LiDARs, they still require training individual networks for each sensor. In contrast, the proposed meta depth completion can handle different sensors with a single network and even generalize to unseen beam patterns.
    A work addressing the challenge of LiDAR-agnostic sparse depth information is Tsai et al.\,\cite{tsai_see_2021}, which, however, focuses on object detection. Their solution is based on applying a ball pivoting algorithm on the point cloud to perform surface completion, yielding a dense surface model. This dense representation serves as the intermediate LiDAR agnostic representation. Lastly, depth information is always sampled in the same manner from the completed shape and passed to the depth completion model. While their concept works for object detection, it is practically infeasible for depth estimation as it would require computing one contiguous surface from all the image pixels. This would be computationally expensive and fail in regions with a low density of depth points.    

\section{Methodology}
    \subsection{LiDAR Inputs}
        The main differentiating characteristics between different LiDAR sensors are \emph{firstly} the number of channels they have \eg how many laser beams are used to scan the environment. \emph{Secondly}, the vertical field-of-view\,(FOV) that the respective sensors have. Both influence how big the vertical angular difference between individual measurements is and, thus, how dense the resulting depth measurements are. For instance, KITTI \cite{geiger_vision_2013} uses a LiDAR with 64 scanning channels while nuScenes\,\cite{holger_caesar_nuscenes_2019} only uses hardware with 32 channels. In other applications, one may prefer sensors with very few channels \cite{single-line:lidar,Voedisch_2022_RAL}. These differences cause significant distribution shifts in the datapoints sampled from the same scenes, posing challenges to state-of-the-art depth completion networks that require dedicated handling. Processing-wise, the concept of LiDAR channels allows fors emulation of lower channel sensors by dropping channels in the data of higher resolution LiDARs allowing the creation of multiple depth images from the same data. \
        
         \begin{figure}
    	    \centering
            \vspace{0.3cm}
    	    \includegraphics[width=0.75\columnwidth]{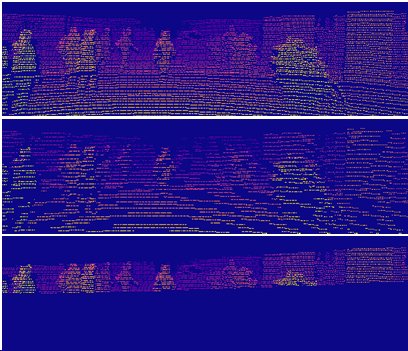}
    	    \caption{\textbf{Sample filtered depth inputs}. Top: Full LiDAR input; Middle: Input after only every fourth channel is retained. The height of the images covers the whole vertical FOV of the LiDAR inputs; Bottom: Reduction of the FOV to a third of the available lines.}
    	    \label{fig:filters}
    	\end{figure}
	
         \subsection{Depth Input Filtering}
    \label{sec:depth_input_filtering}
        Our experiments require the availability of different input styles for the sparse depth data. However, most of the datasets only feature one LiDAR sensor. To train and evaluate our method, we follow \cite{Voedisch_2022_RAL} to synthesise data of different LiDARs by filtering out a certain amount of input channels of a high-end LiDAR. We use the 64-channel LiDAR of KITTI as the high-end LiDAR and have designed two approaches to generate the depth data: \emph{First}, the number of channels is reduced by only considering every $n$-th channel of the original input such that it is sparsified. 
       This resembles a system with the same vertical FOV but a higher angle between adjacent LiDAR channels.
        \emph{Second}, the number of channels is reduced by removing all channels that are outside a specified interval of the vertical field of view $ [ c_{begin},  c_{end}] $. 
        This preserves the distance between adjacent channels but reduces the vertical FOV. The different effects of both filters are visualised in Fig.\,\ref{fig:filters}.
        KITTI does not directly provide the channels of the depth measurements. Hence, they were extracted by dividing the elevation angle of the points evenly, corresponding to the elevation resolution of the Velodyne\,HDL-64E LiDAR sensor used, allowing to assign a channel to each datapoint.

    \subsection{Base Architecture}\label{sec:base-architecture}
        \begin{figure}
    	    \centering
            \vspace{0.3cm}
    	    \includegraphics[width=\columnwidth]{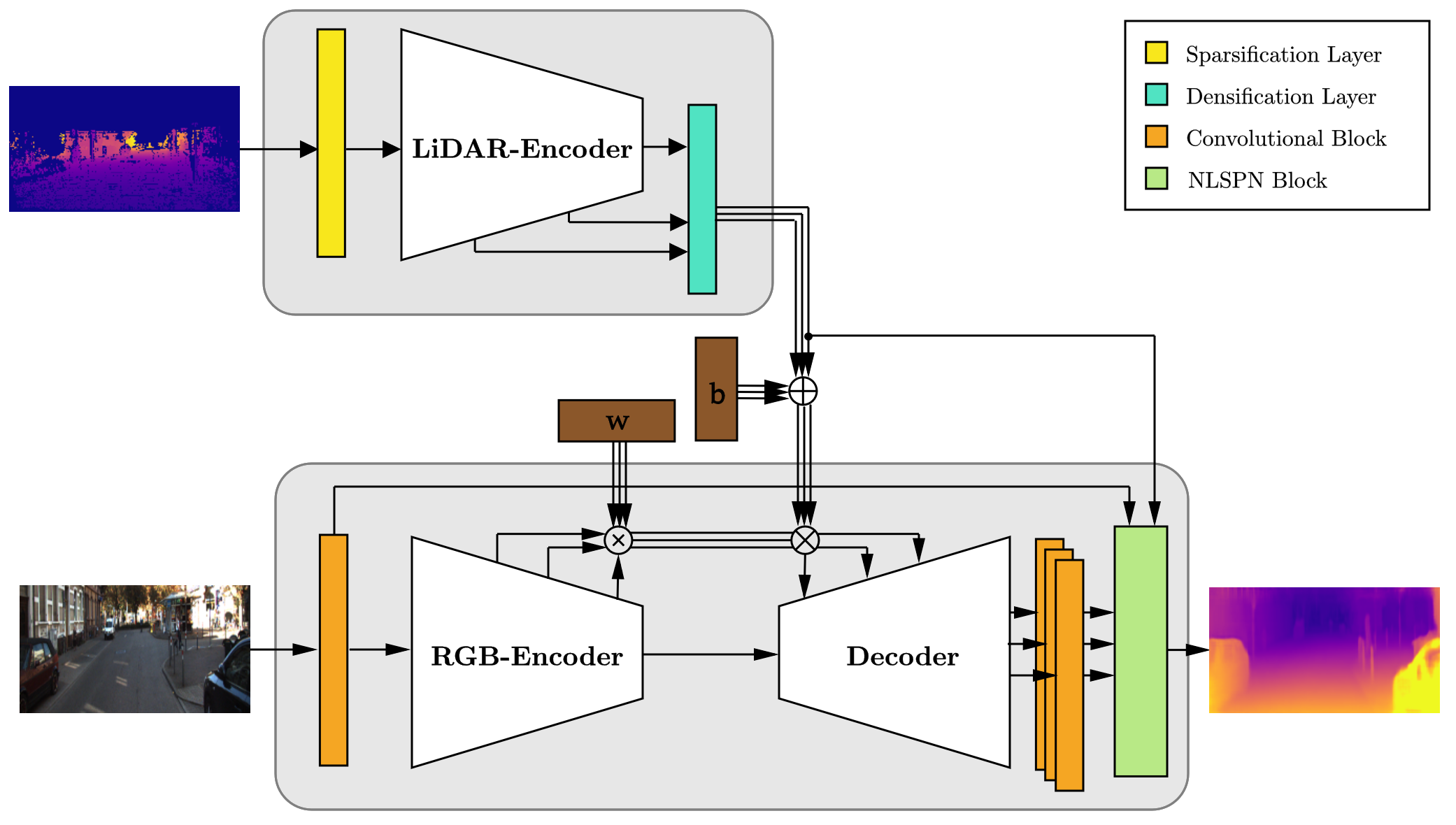}
    	    \caption{\textbf{Basic ResSAN architecture} which is the baseline for the proposed adaption. An NLSPN\,\cite{park_non-local_2020} block was added for better performance. The RGB encoder-decoder consists of ResNet34-inspired blocks. The depth encoder is a sparse encoder adapted from \cite{guizilini_sparse_nodate} that produces a densified output after each layer that is used for the skip connection level fusion of depth and RGB features. $w, b$ are learnable parameters defined in Eq. (\ref{eq:fusion}).}
    	    \label{fig:ResSAN_base}
	    \end{figure}
	    
        Our work utilizes the concept of Sparse-Auxiliary-Networks which were introduced by Guizilini\,\etal\,\cite{guizilini_sparse_nodate} being the first architecture that could perform both monocular depth prediction and full depth input depth completion with the same model. It consists of an encoder-decoder structure for the RGB-data with skip connections and a separate depth encoder for sparse depth input data using a mid-level fusion approach. More precisely, the features from the image and the LiDAR are fused at the skip connections as depicted in Fig.\,\ref{fig:ResSAN_base}. To alleviate the issues encountered with conventional convolutions on sparse data\,\cite{uhrig_sparsity_2017}, the depth encoder is implemented using sparse convolutions\,\cite{choy_4d_2019, yan_revisiting_2020} allowing to neglect spurious information from uninformative areas and also speeding up computations.
        
        From this structure, it follows that the fusion at the skip connections is a promising starting point for a generalization of the SAN architecture for LiDAR-adaptive depth completion. In the following section, we show how the fusion of the features at the skip connection of each abstraction level $i$
        \begin{align}
            \tilde{\mathbf{f}_i} = w_i \cdot \mathbf{f}_i^{\text{rgb}} + \mathbf{f}_i^{\text{lidar}} + b_i
            \label{eq:fusion}
        \end{align}
        can be used to achieve our goal. $\mathbf{f}_i^{\text{rgb}}$ represents the feature map from the RGB inputs while $\mathbf{f}_i^{\text{lidar}}$ is the densified depth feature map. Unlike in the original SAN paper where the fusion weights $w, b$ are trained but fixed at inference time, we use a weight adaption network that calculates fusion weights depending on the input type.
        
        For our implementation we chose to use a ResNet34\,\cite{he_deep_2016} based encoder-decoder structure mainly for reasons of computational cost as opposed to the very big PackNet\,\cite{guizilini_3d_2020} base structure in the reference work. It is from now on referred to as \emph{ResSAN}. Additionally, we include a Non-local Spatial Propagation Network\,(NLSPN)\,\cite{park_non-local_2020} in this architecture to improve the depth predictions. Therefore, the decoder is extended such that it predicts an initial depth estimate, affinity matrices and a depth confidence matrix which is then used by the NLSPN to refine the depth prediction. This module allows us to obtain state-of-the-art predictions with comparably small encoder-decoder structures.

        \begin{figure}
    	    \centering
            \vspace{0.3cm}
    	    \includegraphics[width=0.95\columnwidth]{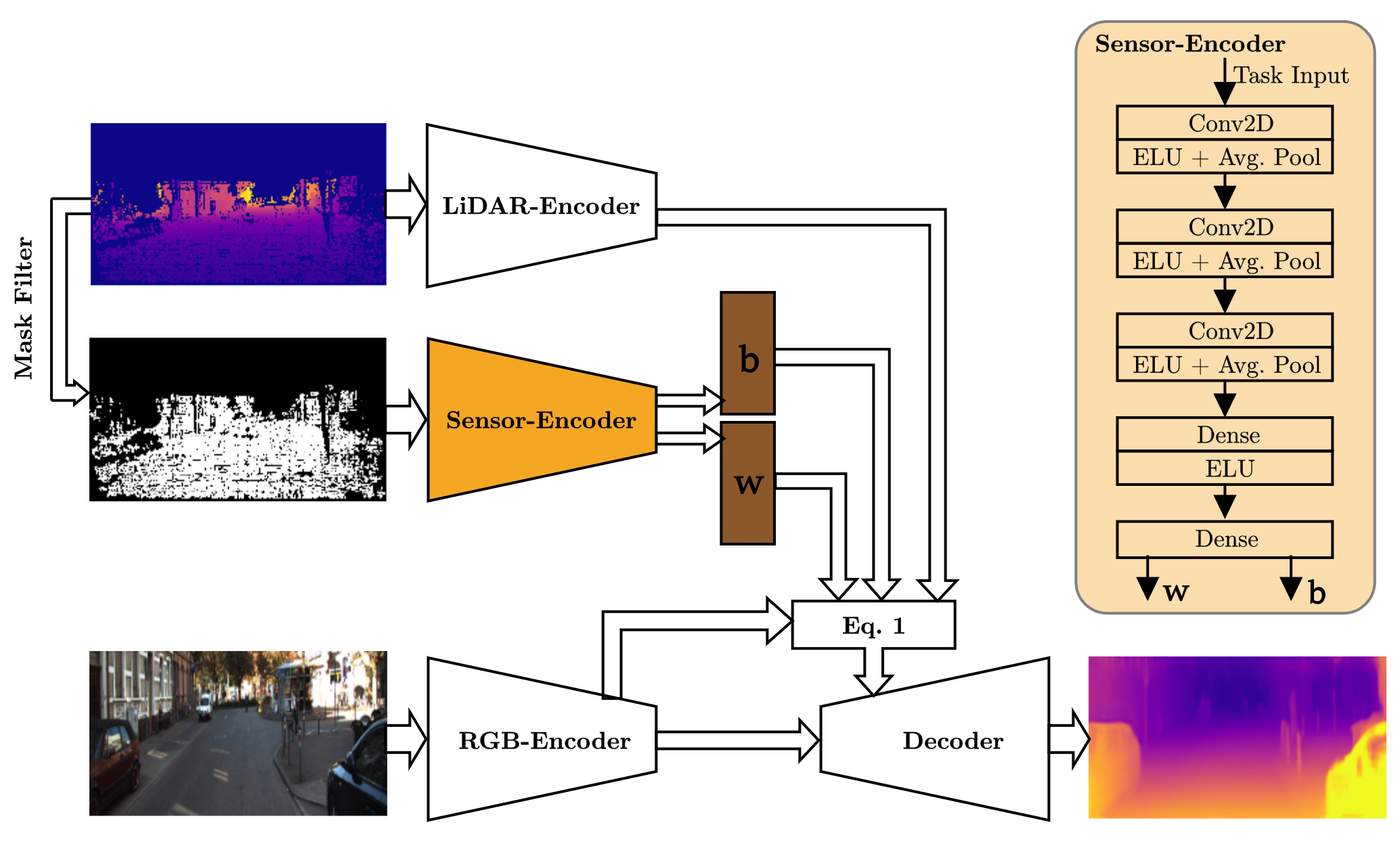}
    	    \caption{\textbf{Our Residual LiDAR Adaptive Network}. All fusion weights are determined by a Sensor-Encoder network and are therefore variable and no longer constant. The task input is directly derived from the depth input through a binary mask of the depth input.}
    	    \label{fig:ResSAN_implicit}
    	\end{figure}

    \subsection{LiDAR Adaptive Network}
        Our approach to extending the SAN concept is to view the task of feeding different kinds of LiDAR depth maps to the network as a multitask-like job. Specifically, each type of depth input that is to be put into the network is considered one task. Therefore, the network needs to have a third input besides RGB-image and depth image, which can be leveraged to infer the depth type of the current batch via a sensor encoder. This setting allows applying the ideas presented in \cite{sun_task_nodate}, namely that we do not have to have one specialized network for every single task but instead use one network with adaptive weights. In the case of the SAN architecture, the weights of the fusion module can be made adaptive as they control the fusion of LiDAR and RGB-related feature maps. The original work on SAN\,\cite{guizilini_sparse_nodate} already demonstrated that the decoder can be trained to work on two input settings properly. Therefore, it is expected that the decoder in this architecture can be trained to work with inputs that are in between the two extremes, no depth input and full LiDAR input, as well.
        
        One possible approach for implementation could be to use a task vector that explicitly specifies the LiDAR scanning pattern for the network. However, such a strategy would inherently fix the possible input style for depth at training time and therefore limit the general usability of the model. Consequently, a generalised encoding of the depth input type is favourable, seeing the space of possible input types more as a continuous space rather than discrete. Therefore, we derive the input for the sensor encoder directly from the input depth map. Applying this approach alleviates the problem of explicitly designing a suitable input space instead of allowing the sensor encoder network to extract the necessary information. Given that the fusion weights need to be adapted to differences in the density of depth values and/or the spatial distribution of the data points, the depth input image is preprocessed such that characteristics can still be extracted without providing too much additional information, which could adversely affect the training. Therefore, we decided to classify the pixels of the depth input 
    	\begin{align}
    		m(p) = \left\{
    		\begin{array}{ll}
    			1  & \mbox{if } p > 0, \\
    			0 & \mbox{otherwise } 
    		\end{array}
    		\right.
    	\end{align}
    	depending on whether they contain depth information or not with $p$ being the depth value of the pixel and $m(p)$ the value of the same pixel in the resulting depth mask. The result is a binary mask that does not contain any depth information but retains the characteristics of the density and spatial distribution of the pixels that contain depth information.
    	
    	This representation is subsequently passed through the sensor encoder, which consists of convolution blocks followed by a dense block as depicted in Fig.\,\ref{fig:ResSAN_implicit}. As we are operating on a binary input map and are interested in density information, the Average Pooling operation was chosen over the more commonplace MaxPooling. The output of the last convolutional layer is flattened, and passed to two dense layers, with the last one split into two parallel layers, one for each output vector\,($w, b$) that constitute the fusion weights. As introduced in (\ref{sec:base-architecture}) the last layer has a linear activation to allow for a sufficiently large value range of the fusion weights. The fusion weights predicted by the task encoder are then used like in the baseline architecture. The resulting overall architecture is from now on referred to as the \emph{Residual Lidar Adaptive Network\,(ResLAN)}.

\section{Experimental Procedure}
    \paragraph{Dataset}
        We train and evaluate our method using the KITTI depth completion dataset\,\cite{geiger_are_2012} with the Eigen split\,\cite{eigen_depth_2014}. Following common practice\,\cite{lee_big_2021}, we apply color jitter, random crop and horizontal flipping to the training inputs. Moreover, the input images are down-sampled to 640x192 pixels to reduce computational load.

	\paragraph{Implementation details}
	    We train our models using PyTorch on a single Nvidia Titan Xp GPU with a batch size of 8, using the Adam\,\cite{kingma_adam_2017} optimizer with a learning rate of $10^{-4}$, $\beta_1 = 0.9$, $\beta_2 = 0.999$ and decay of $10^{-2}$. We apply  SILoss\,\cite{eigen_depth_2014}, and for runs including multiple styles of depth input, all types were applied with the same loss weight and number of samples shown. The training schedule is as follows: For the first 10 epochs, we train our model with RGB images only, with the depth encoder, weight adaption and NLSPN being disabled. Then, we fix the RGB-encoder and resume training the complete model using depth inputs for another 10 epochs.
    
\section{Experimental Results}\label{sec:results}
    \subsection{General Results}
        The basic ResSAN model is used to determine reference results which our expanded model can be compared to as it is structurally similar to ResLAN but does not possess the Lidar adaptive components of it. Further, we compare with the full-size PackNet-SAN and the unmodified NLSPN architecture. 
        As it can be seen from Tab.\,\ref{tab:sota-results}, our LiDAR-adaptive ResLAN achieves competitive performance compared to state-of-the-art standard depth completion methods, which are specialized to the unfiltered 64-beam-LiDAR. The performance differences are in the range of a few centimetres in terms of MAE, which is acceptable given the practical advantage that ResLAN can generalize to different beam patterns as will be shown below.

\begin{figure}
            \centering
            \vspace{0.1cm}
            \includegraphics[width=0.95\columnwidth]{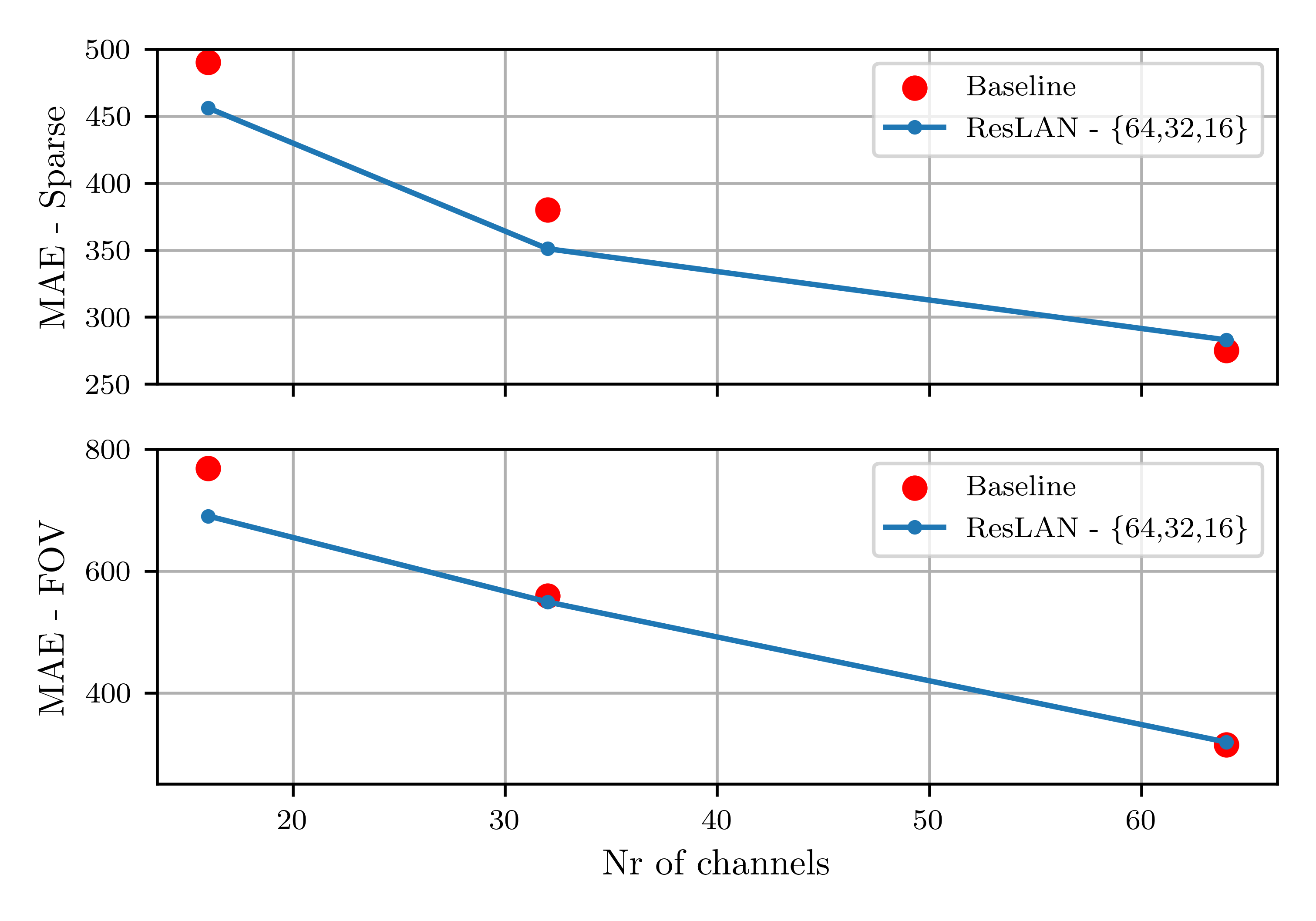}
            \vspace{-0.1cm}
            \caption{\textbf{Comparison of ResLAN models with ResSAN} baseline models which were specifically trained for one depth input type as opposed to the ResLAN models.}
            \label{fig:comp-plot}
            \vspace{-0.1cm}
        \end{figure}

        Furthermore, we compared the architectures for a set of three different input types that contained 64, 32 or 16 LiDAR channels using both filter types on the metrics from the KITTI benchmark. The NLSPN model was trained for the standard depth completion task and then evaluated with different input data. As for the ResSAN models, we trained one model for each input type and tested it for the corresponding one which serve serve as the \emph{Baseline} in Tab.\,\ref{tab:overall-results}. Our ResLAN model was jointly trained for all three settings. As listed in Tab.\,\ref{tab:overall-results}, the ResLAN models outperform the challenging baseline in all metrics for FOV filtering and all but one for sparse filtering. This implies that our LiDAR adaptive model is able to outperform dedicated models in case of very sparse input depth. Fig.\,\ref{fig:comp-plot} shows this is indeed the case for 32 and even more for 16 channels. For FOV-filtered inputs with 16 channels, the ResLAN exhibits approx. $10\%$ smaller MAE than the baseline. As for the NLSPN, it becomes apparent that it is not capable of generalizing to other input types since it shows clearly worse results. The difference is especially pronounced for the FOV filtering where on average more than every fourth predicted pixel is more than $25 \%$ deviating from the ground truth\,($\delta_{1.25}$). Therefore, using a weight-adapting network in combination with differently filtered input depths allows us to train models that outperform their non-adaptive counterparts.

        \begin{table}[]
            \centering
    	    \small
            \vspace{0.4cm}
            \caption{\textbf{Depth estimation result for standard depth completion} when the ResSAN model was only trained for 64 channels and the ResLAN model for multiple tasks. The PackNet-SAN and NLSPN models were trained with the setup that was also used for our model architecture.}
            \footnotesize
            \setlength{\tabcolsep}{5pt}
            \begin{tabular}{@{}lrrrrl@{}}
            \toprule
            \multicolumn{6}{c}{\textbf{Standard LiDAR Depth Completion}}                                                                                                                         \\ \midrule
            \multicolumn{1}{l|}{Method}          & RMSE $\downarrow$            & MAE  $\downarrow$            & iRMSE $\downarrow$             & iMAE $\downarrow$ & $\delta_{1.25}$ $\uparrow$ \\
            \multicolumn{1}{l|}{}                & \multicolumn{1}{l}{{[}mm{]}} & \multicolumn{1}{l}{{[}mm{]}} & \multicolumn{1}{l}{{[}1/km{]}} & {[}1/km{]}        &                            \\ \midrule
            \multicolumn{1}{l|}{PackNet-SAN}     &  914                            &  298                            &  2.78                              &  1.4                 &  99.65 \%                          \\
            \multicolumn{1}{l|}{NLSPN}           &  \textbf{889}                            &   \textbf{263}                           &  \textbf{2.62}                              &   \textbf{1.3}                &   \textbf{99.61} \%                         \\ \midrule
            \multicolumn{1}{l|}{ResSAN (Ours)}   & 948                             &  275                            &  2.75                              &    1.4               &   99.58 \%                         \\
            \multicolumn{1}{l|}{ResLAN (Ours)} &   969                           &  283                            &   2.83                             &   1.4                &  99.56 \%                          \\ \bottomrule
            \end{tabular}
            \vspace{0.2cm}
            \label{tab:sota-results}
        \end{table}

        \begin{table}[]
    	    \centering
    	    \small
    	    \caption{\textbf{Depth estimation results of the two baseline setups and the explicit and implicit ResSAN} when evaluated on a combination of 16, 32 and 64 channel depth inputs. Please note that Specialist Methods need to train three specialized networks, one for each of the three types of inputs while our method only uses one network.}
            \footnotesize
            \setlength{\tabcolsep}{4.8pt}
            \begin{tabular}{@{}lrrrrl@{}}
                \toprule
                \multicolumn{6}{c}{\textbf{Sparse Channel Filter}}                                                                                                                                  \\ \midrule
                \multicolumn{1}{l|}{Method}        & RMSE $\downarrow$            & MAE  $\downarrow$            & iRMSE $\downarrow$             & iMAE $\downarrow$ & $\delta_{1.25}$ $\uparrow$  \\
                \multicolumn{1}{l|}{}              & \multicolumn{1}{l}{{[}mm{]}} & \multicolumn{1}{l}{{[}mm{]}} & \multicolumn{1}{l}{{[}1/km{]}} & {[}1/km{]}        &                             \\ \midrule
                \multicolumn{1}{l|}{NLSPN}         &  1396                            &  437                            & 5.54                               &  2.2                 &  98.82 \%                           \\
                \multicolumn{1}{l|}{Baseline}      & \textbf{1207}                             &  381                            & 4.41                               &  1.8                 &  \textbf{99.37} \%                           \\
                \multicolumn{1}{l|}{ResLAN (Ours)} &  1215                            &  \textbf{378}                            &  \textbf{4.27}                              &  \textbf{1.7}                 &  99.31 \%                           \\ \toprule
                \multicolumn{6}{c}{\textbf{Field-of-View Filter}}                                                                                                                                   \\ \midrule
                \multicolumn{1}{l|}{Method}        & RMSE $\downarrow$            & MAE  $\downarrow$            & iRMSE $\downarrow$             & iMAE $\downarrow$ & $\delta_{1.25}$ $\uparrow$ \\
                \multicolumn{1}{l|}{}              & \multicolumn{1}{l}{{[}mm{]}} & \multicolumn{1}{l}{{[}mm{]}} & \multicolumn{1}{l}{{[}1/km{]}} & {[}1/km{]}        &                             \\ \midrule
                \multicolumn{1}{l|}{NLSPN}         &  2738                            &  1702                            & 12.3                              &  4.3                 &  74.69 \%                           \\
                \multicolumn{1}{l|}{Baseline}      &  1556                            &  525                            &  6.8                              &  3.0                 & 98.14 \%                            \\
                \multicolumn{1}{l|}{ResLAN (Ours)} &  \textbf{1548}                            &  \textbf{519}                            &  \textbf{6.44}                              &  \textbf{2.8}                 & \textbf{98.52 \%}                            \\ \bottomrule
            \end{tabular}
            \label{tab:overall-results}
        \end{table}

        \begin{figure*}
    	    \centering
    	    \includegraphics[width=0.8\textwidth]{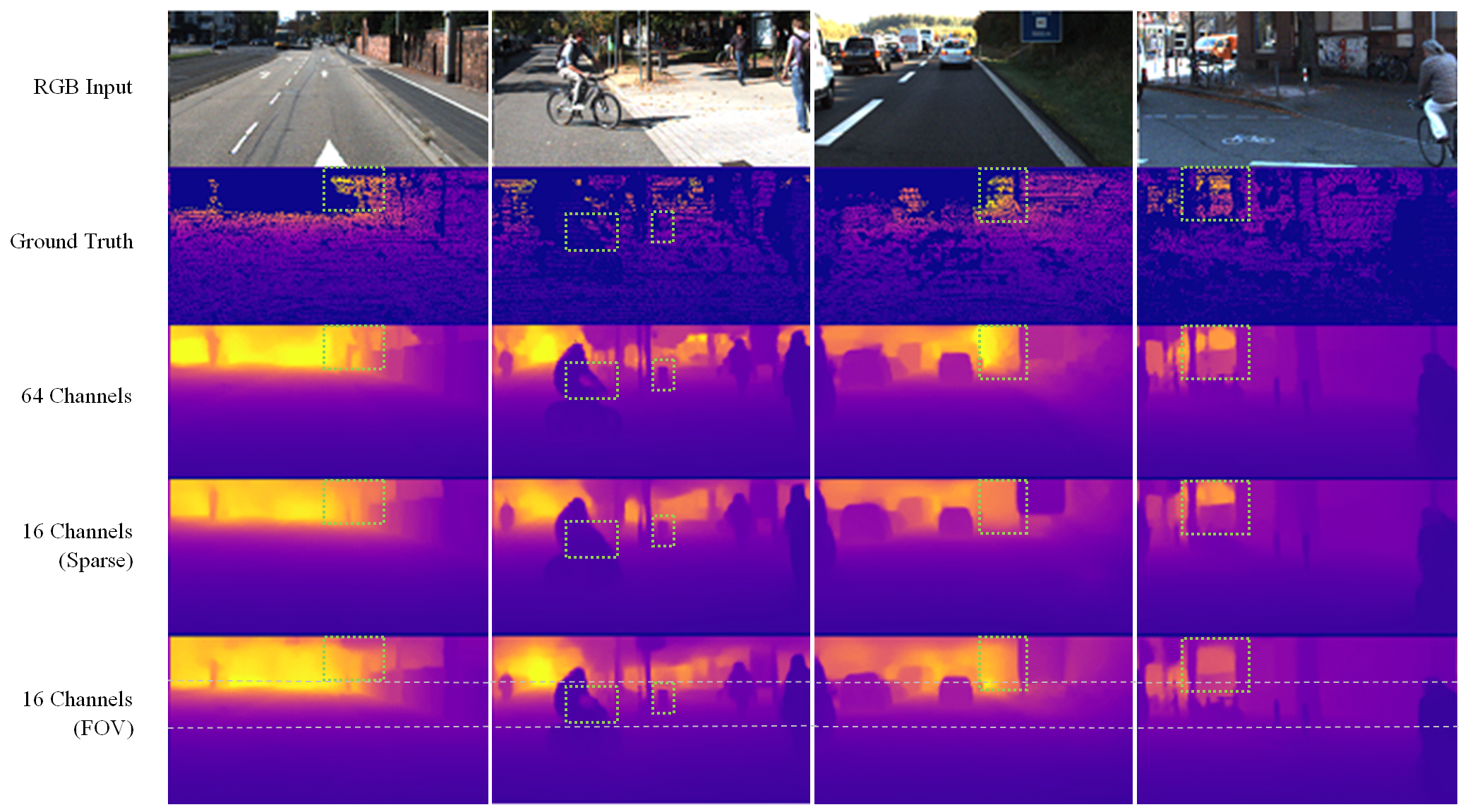}
    	    \caption{\textbf{Depth maps} for four different sample inputs and three different depth filters returned by a single ResLAN model. Green dotted boxes highlight areas where the models behave clearly differently. The grey dotted line visualises the area of the image where depth data was available. Note that the depth maps within the grey dotted boundaries are about as good as for the 64 channel LiDAR input case while areas with no depth info have lower quality.}
    	    \label{fig:dense-maps}
            \squeezeup
    	\end{figure*}

    \subsection{Filter Effects}
        Comparing the effect of the two different types of depth input filters on the model performance, it becomes apparent that FOV filtering is the more challenging task. In that setting, reducing LiDAR channels is more detrimental to the performance than sparse filtering as it creates regions where no depth information is available. Effectively, the model is forced to perform depth prediction in these regions. These effects are highlighted in the depth images in Fig.\,\ref{fig:dense-maps} where the effect of a 16-channel sparse depth filter and a 16-channel FOV can be compared.

    \subsection{Generalization Capabilities}
\begin{figure}
        \centering
        \vspace{-0.3cm}
        \includegraphics[width=0.9\columnwidth]{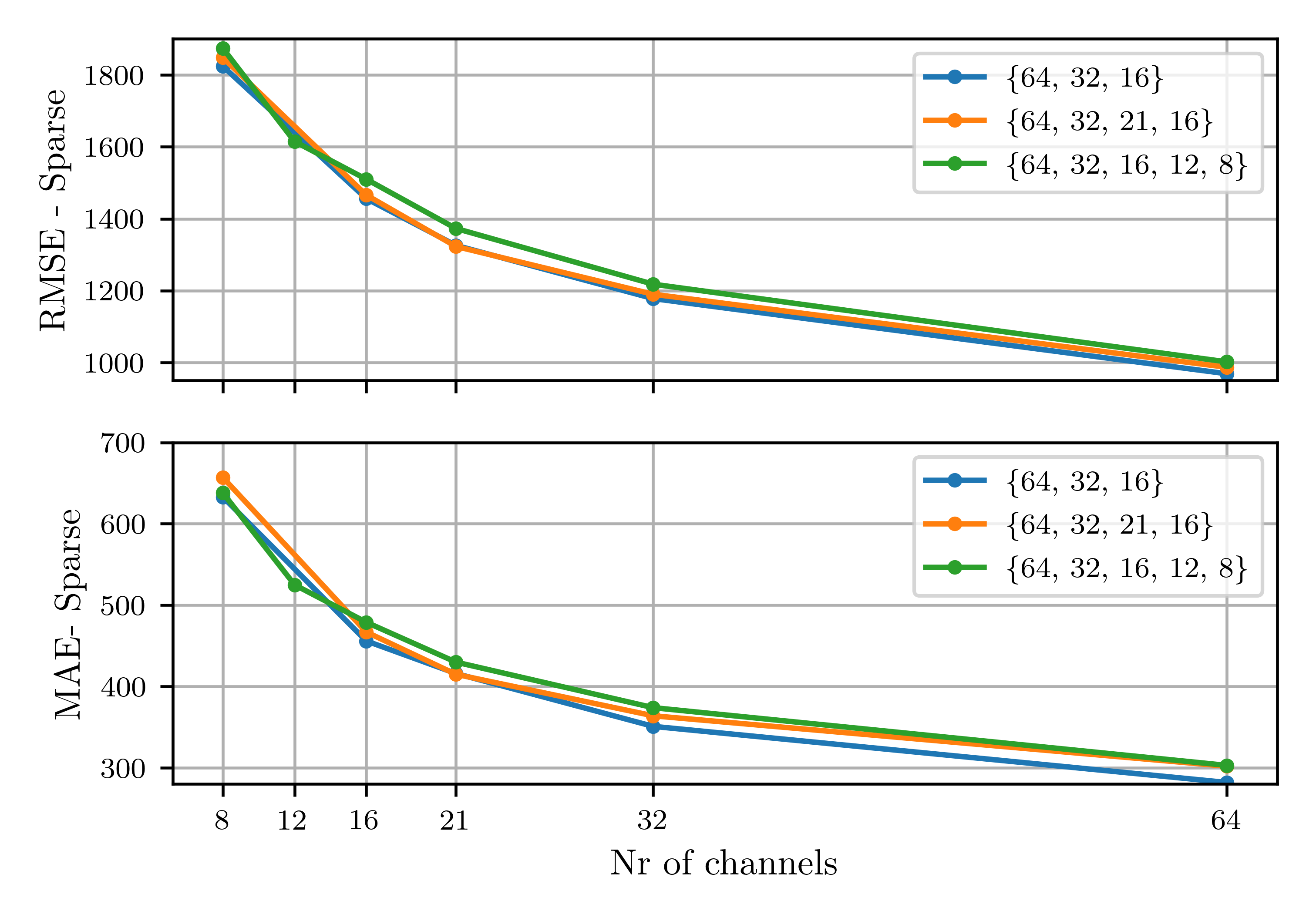}
        \vspace{-0.3cm}
        \caption{\textbf{ResLAN models with sparse channel filtering} - One line represents one model and a datapoint the RMSE respectively MAE loss for that particular number of depth input channels. The channel numbers the models were trained for are given in the legend.  All other datapoints represent cases where the model had to generalise.}
        \label{fig:explicit-comp}
        \squeezeup
    \end{figure}

\begin{figure}
        \centering
        \vspace{-0.3cm}
        \includegraphics[width=0.9\columnwidth]{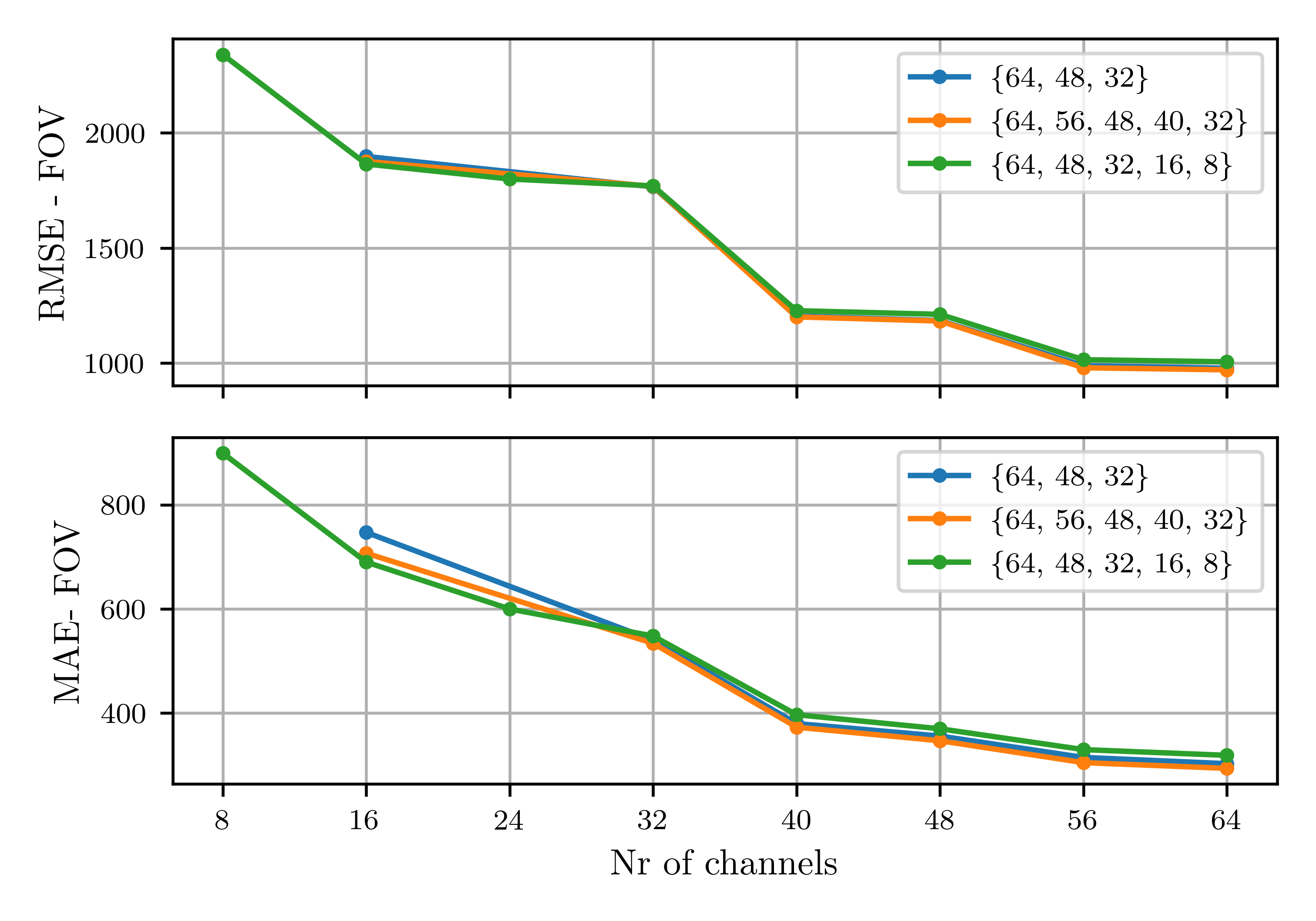}
        \vspace{-0.3cm}
        \caption{\textbf{ResLAN models with Field-of-View filtering} - One line represents one model and a datapoint the RMSE respectively MAE loss for that particular number of depth input channels. The channel numbers the models were trained for are given in the legend. All other datapoints represent cases where the model had to generalise.}
        \label{fig:implicit-comp}
        \squeezeup
    \end{figure}

        We trained three models for both filter types eaach, so the combinations and number of filtered depth inputs they receive are different. This serves the purpose of testing the generalization capabilities of the ResLAN architecture as well as the robustness to different filter settings. After training, the models were evaluated for the depth input settings they were trained for, as well as for ones they weren't exposed to. Overall, ResLAN shows good generalization capabilities. As one can gather from Fig.\,\ref{fig:explicit-comp} and Fig.\,\ref{fig:implicit-comp}, the consequences of slightly varying sets of input depth settings are limited. The most considerable deviations can be seen when the model is tasked to extrapolate. For instance, the model $\{64, 32, 16\}$ shows a noticeably higher MAE for eight-channel depth inputs than the model that was trained for it. Similar behaviour can be seen for the FOV filtering case as well for the model $\{64, 48, 32\}$ when tasked to generalize for a 16-channel input. There is no such pronounced effect for generalization tasks that lie between two filter settings the model was trained for. At most, it can be observed that models that were trained for a smaller range of filter values perform slightly better than ones that have to cover a wider range. The number of filter settings used in a fixed range does not relevantly influence the model performance, as can be seen, when comparing the two models in Fig.\,\ref{fig:implicit-comp}, which are both trained for a range of 64 to 32 channels but one with three filter settings and the other one with five.
    
\section{Conclusions}
    This paper presents a novel LiDAR-adaptive meta depth completion method that allows the training of a single depth completion model capable of operating on different LiDAR input patterns and without sparse depth inputs. It uses an adaptive middle-fusion approach on skip connections and distinguishes itself by its capability to generalise to previously unseen input styles. Moreover, our experiments have shown that the joint training for multiple types of sparse depth inputs allows for the models to be better for very sparse inputs than models trained only for this specifically. Hence, meta depth completion can also be beneficial for certain cases of single input type depth completion, especially for cases where only very sparse depth data is available.  In future work, we will extend the synthetic filtering to mimic other real LiDAR inputs to evolve the training quality for our meta depth learning. Moreover, we will explore approaches to further develop our model architecture to close the accuracy gap for full LiDAR input of the ResLAN to the state-of-the-art standard depth completion architectures.

{\small
\bibliographystyle{ieee_fullname}
\bibliography{references}   
}

\newpage
\appendix
\section{Supplement}
\subsection{Overview}
    The supplementary material elucidates further on the filtering of the LiDAR depth input in Sec.\,\ref{app:depth} and a more detailed explanation of the experimental procedure\,\ref{app:proc}.
\subsection{Depth Data Filtering}\label{app:depth}
	For our experiments, we have designed two classes of augmented LiDAR data. \emph{Firstly}, the number of channels is reduced by only considering every $n$-th channel of the original input. This resembles a system with the same vertical FOV but a higher angle between adjacent LiDAR channels. From now on, this filter will be referred to as sparse channel filter for briefness' sake. \emph{Secondly}, the number of channels is reduced by removing all channels that are outside a specific interval, therefore, keeping the angle between adjacent channels constant but reducing the vertical FOV.\\

	\paragraph{Channel identification}
	The datapoints in the depth input image do not contain any explicit information on the channel (i. e. the laser beam that captured them), making it necessary to process the depth map in a way that a LiDAR channel can be assigned to each data point. For this, we use the property of the Velodyne that the vertical angle\,$\theta$ of each laser can be assumed to be nearly constant. Additionally, the angular difference between two adjacent lasers is constant for the used device. Based on this, it is possible to map each channel to a specific vertical angle interval. Using the characteristics specified for the Velodyne 
	{\small
	\begin{align}
		\theta_{\min} & =-24.9^\circ \\
		\theta_{\max} &= +2.0^\circ \\
		\Delta \theta &= \theta_i - \theta_{i-1} = \text{const.}, \hspace{1.5cm} i \in 0,...,63
	\end{align}%
	}%
	the channel 
	{\small
	\begin{align}
		c(\theta) = \text{round}\left( (\theta - \theta_{\min}) \cdot \frac{63.99}{\theta_{\max}-\theta_{\min}} - 0.5 \right)
	\end{align}%
	}%
	is estimated using the aforementioned considerations utilizing the rounding operation as the classifier that returns an integer class label in the range of $0,1,..,63$. $\theta$ represents the vertical angle with respect to a horizontal line, $\theta_{\min}$ is the minimum vertical angle and $\theta_{\max}$ is the maximum one.\\
	
	\paragraph{Re-projection into 3D space}
	The aforementioned estimation algorithm to generate a channel label for each datapoint requires that $\theta$ is known, which is not initially the case. The positional data of each depth point is compressed to a Cartesian pixel coordinate in the image. Extracting the vertical angle requires projecting each datapoint back into 3D space. Such a projection from 2D to 3D space would normally be an ill-posed problem. However, as we have the depth information on each valid pixel, the problem has a unique well-posed solution in our case. From the calibration matrix of the camera where skew coefficient $s=0$, one can obtain the parameters necessary to perform the derivations
	{\small
	\begin{align}
		\frac{X}{Z} &= \frac{x - u_0}{fk_x}\\
		\frac{Y}{Z} &= \frac{y - v_0}{fk_y}\\
		d &= \sqrt{X^2 + Y^2 + Z^2}
	\end{align}%
	}%
	taken from the geometrical image formation formulas and the Euclidean distance in 3D. $(x,y)$ are the pixel coordinates of the depth point, $(X,Y,Z)$ the corresponding coordinates in 3D space, $fk_x, fk_y$ the focal lengths in terms of pixels and ($u_0$, $v_0$) the principal point.
	Having the value of $Y$ allows for the calculation of the vertical angle
	{\small
	\begin{align}
		\theta = \arcsin \left(\frac{Y}{d + \epsilon} \right)
	\end{align}%
	}%
	with $\epsilon = 10^{-5}$ being a modification for computational reasons to avoid NaN in case of depth values of zero allowing to process the whole image and filter out the invalid pixels later.
	
	\paragraph{Filtering}
	\begin{figure}
	    \centering
        \vspace{0.5cm}
	    \includegraphics[width=\columnwidth]{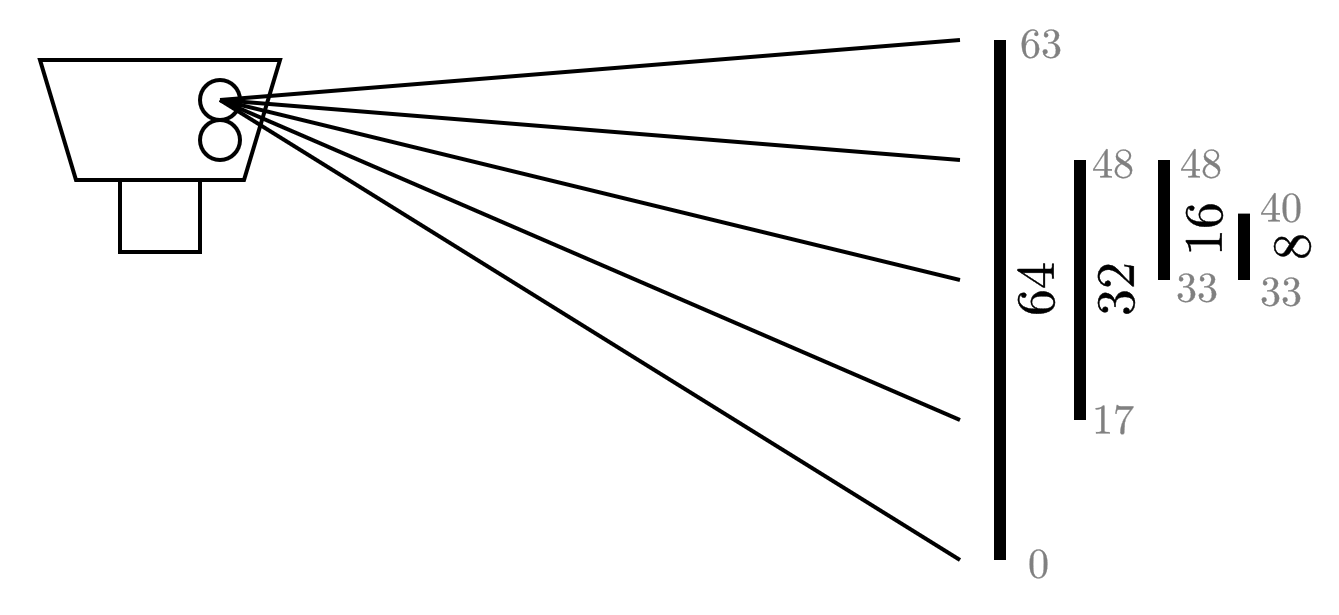}
	    \caption{Illustration of a selection of the filters for vertical FOV reduction. The grey numbers indicate from which channel to which channel each filter is defined. The filters are always referred to by the number of channels they retain.}
	    \label{fig:FOV-vis}
	\end{figure}
	
	After assigning a LiDAR channel to each datapoint in the depth image, it is now possible to perform the filtering. For the first filter method which keeps every $n$-th channel, a datapoint is kept if its channel satisfies
	{\small
	\begin{align}
		0 = c(\theta(p)) \mod n \hspace{0.4cm},
	\end{align}%
	}%
	with $c(\theta(p))$ being the channel of the individual pixel $p$ that has the vertical angle $\theta(p)$. Datapoints which do not satisfy this condition are discarded and replaced with zero values.
	The second method keeps all datapoints that have a channel within
	{\small
	\begin{align}
		c(\theta(p)) \in [ c_{begin},  c_{end}]
	\end{align}%
	}%
	with $ c_{begin}$ being the lowest channel and $c_{end}$ being the highest channel that is kept. The precise settings used in the experiments are illustrated in Fig.\,\ref{fig:FOV-vis} and provided in detail in Tab.\,\ref{tab:FOV-vis}.

    \begin{table}[tbh]
    \centering
        \caption{Starting and End channels for the different filter settings used for the Field-of-View filter.}
        \begin{tabular}{@{}ccc@{}}
            \toprule
            \multicolumn{3}{c}{\textbf{Field-of-View Filter}} \\ \midrule
            Nr. channels    & $ch_{begin}$    & $ch_{end}$\    \\ \midrule
            64              & 0               & 63            \\
            56              & 7               & 63            \\
            48              & 7               & 55            \\
            32              & 17              & 48            \\
            24              & 25              & 48            \\
            16              & 33              & 48            \\
            8               & 33              & 40            \\ \bottomrule
        \end{tabular}
        \label{tab:FOV-vis}
    \end{table}

\newpage
\subsection{Experimental Procedure}\label{app:proc}

This section illustrates further details of the training procedure used to generate the results in Sec.\,\ref{sec:results} and also further analysis of the latter.\\

\paragraph{Sparse Channel Filter}
For the sparse channel filter, all models were evaluated for filter settings that retain 64, 32, 21, 16 and 8 scan lines of the LiDAR sensor. Two models were trained for channels ranging from 64 to 16 channels. One was trained for 64, 32, and 16 channels while the other was also trained for 21 channels. This allows us to compare the effect of the number of filter settings on the performance. As shown in Fig.\,\ref{fig:explicit-comp} this has minuscule consequences. Moreover, a third model was trained on a channel range from 64 to 8 channels with training for 64, 32, 16, 12 and 8 channels focusing more on cases of very sparse LiDAR data. It can be noted that extending the range of possible sparsity levels is not beneficial to overall performance since this method suffers from slight performance degradation for all filter settings except for the 8-channel case where the other models needed to extrapolate.\\

\paragraph{Field-of-View Filter}
For the FOV filter, we likewise trained two models on the same range of retained scan lines. Both were evaluated for 16, 32, 40, 48, 56 and 64 channels. One of the models was trained for 64, 48, and 32 channels. The other one was moreover also trained for data that retained 40 and 48 channels reducing the difference in channels between the filter settings. Especially for the MAE metric we can observe that the model that was trained for more FOV filter settings performs slightly better as can be seen from Fig.\,\ref{fig:implicit-comp}. Noteworthy is that it also exhibits a better generalisation capability when presented with 16-channel data, which it was not trained for. The third model was trained for 5 different channel filters but this time on the range from 64 channels to 8 channels. As previously seen with the sparse channel filter, extending the range of filter settings for the FOV entails small overall performance losses.

\addtolength{\textheight}{-12cm}   %

\end{document}